\documentclass{article}
\usepackage{M-PSI}

\usepackage{lipsum} 
\usepackage{graphicx}
\usepackage{soul}
\usepackage{multirow}
\usepackage{rotating}
\usepackage{amssymb}

\removefirstpagelogo{}

\hypersetup{
  hidelinks,
  urlcolor=red,
  pdftitle={Multimodal Group Emotion Recognition In-the-wild Using Privacy-Compliant Features},
  pdfauthor={Anderson Augusma, Dominique Vaufreydaz, Fr\'ed\'erique Letu\'e},
  pdfkeywords={Multimodal, Privacy safe, Transformer networks, Group emotion recognition in-the-wild.},
}

\title{Multimodal Group Emotion Recognition In-the-wild Using Privacy-Compliant Features}

\author{\href{https://www.m-psi.fr/augusma-anderson/}{Anderson Augusma\textsuperscript{\tinyspace{}1,2,~\small{\ExternalLink}}}, 
\href{https://research.vaufreydaz.org/}{Dominique Vaufreydaz\textsuperscript{\tinyspace{}1,~\small{\ExternalLink}}}, 
\href{https://membres-ljk.imag.fr/Frederique.Letue/}{Fr\'ed\'erique Letu\'e\textsuperscript{\tinyspace{}2,~\small{\ExternalLink}}}\vspace{0.1cm}\\
{\textsuperscript{1} \href{https://www.m-psi.fr/}{Univ. Grenoble Alpes, CNRS, Grenoble INP, LIG, 38000 Grenoble, France\textsuperscript{\tinyspace{}\small{\ExternalLink}}}}\\ %
{\textsuperscript{1} \href{https://svh.imag.fr}{Univ. Grenoble Alpes, CNRS, LJK, 38000 Grenoble, France\textsuperscript{\tinyspace{}\small{\ExternalLink}}}}\\
}

\renewcommand{\subtitleFigure}
{
    \centering
    \captionsetup{type=figure}
    \includegraphics[width=\linewidth]{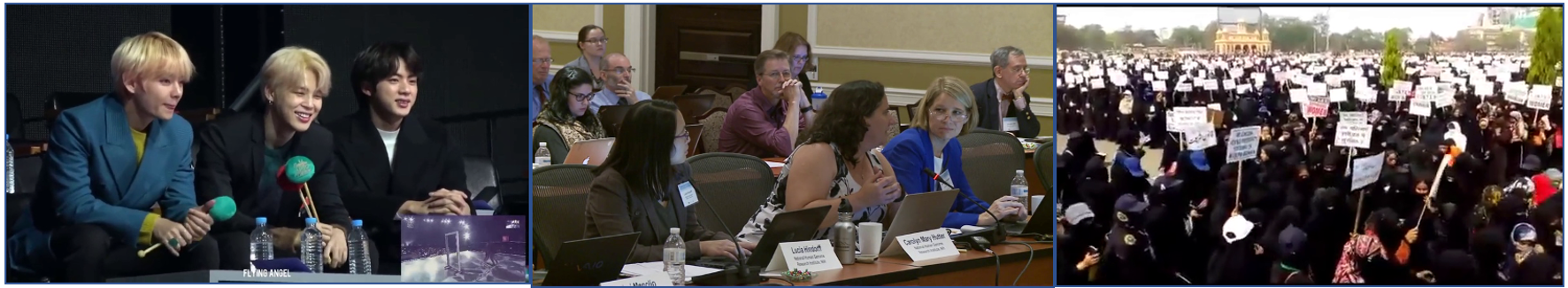}
    \captionof{figure}{Examples from the VGAF dataset. From left to right, examples of Positive, Neutral, and Negative classes.}
    \label{fig:example}
    \vspace{0.5cm}
}

\begin{document}

\setcounter{footnote}{0}

\begin{abstract}
This paper explores privacy-compliant group-level emotion recognition "\textit{in-the-wild}" within the EmotiW Challenge 2023.
Group-level emotion recognition can be useful in many fields including social robotics, conversational agents, e-coaching and learning analytics.
This research imposes itself using only global features avoiding individual ones, i.e. all features that can be used to identify or track people in videos (facial landmarks, body poses,  audio diarization, etc.). 
The proposed multimodal model is composed of a video and an audio branches with a cross-attention between modalities.
The video branch is based on a fine-tuned ViT architecture.
The audio branch extracts Mel-spectrograms and feed them through CNN blocks into a transformer encoder.
Our training paradigm includes a generated synthetic dataset to increase the sensitivity of our model on facial expression within the image in a data-driven way.
The extensive experiments show the significance of our methodology. 
Our privacy-compliant proposal performs fairly on the EmotiW challenge, with 79.24\% and 75.13\% of accuracy respectively on validation and test set for the best models.
Noticeably, our findings highlight that it is possible to reach this accuracy level with privacy-compliant features using only 5 frames uniformly distributed on the video.

\end{abstract}

\vfill{}
\section{Introduction}

Emotion recognition research is of interest in  multimodal interaction for numerous applications like social robotics, conversational agents, e-coaching, or learning analytics.
Among others, one challenge is the automatic recognition of group-level emotions in ecological or "\textit{in-the-wild}" scenarios, which involves considering group dynamics, individual behavior, emotional expressions, postures, environmental elements, and different kinds of activities~\cite{Sharma2019}.
This task is further complicated by cultural and ethnic factors and some technical issues like the quality of recording point of view.
Adding concerns about ethics and privacy limits some choices in the data that can be employed as input and outputs of the machine learning algorithms. 

This article presents a privacy-compliant architecture to classify audio-visual data labeled with group-level emotions: the VGAF dataset~\cite{Sharma2019}.
This machine learning model materializes our participation in the EmotiW 2023~\cite{emotiw2023} challenge.
As our former research, Petrova et al.~\cite{Petrova20}, we impose ourselves some privacy rules on the input of our machine learning model.
Unlike other approaches~\cite{savchenko2022neural,sharma2021audio,Sun2020,Liu2020}, our privacy-compliant model must not use any information that can be helpful to identify or track any person on the videos. 
It avoids individual data like postures or facial landmarks, for instance, focusing only on global features computed on images and sound.

The paper is structured as follows. After the presentation of related work (section~2) and the VGAF dataset (section~3), section~4 introduces our methodology. Section~5 details our extensive experiments while section~6 discusses the results in regard to the state-of-the-art.

\section{Related Work}

Progresses in machine learning in the last decade permit to address emotion recognition in real-world  or ``\textit{in the wild}'' scenarios, and more recently group level emotion prediction in videos.
While some research remains monomodal~\cite{Petrova20,ottl2020group,savchenko2022neural}, most of them use multimodal inputs~\cite{Evtodienko2021,pinto2020audiovisual,belova2022group}.
Surrace et al.~\cite{Surace2017} used a method that examines both general and specific information.
They focused on the overall scene in a top-down strategy.
They also extracted personal information in a bottom-up strategy, isolating faces from wider images.
Other studies~\cite{Guo2018, belova2022group, savchenko2022neural, sharma2021audio, Guo2020, Gupta2018, Liu2020, Yu2019, Hossain2019, Wang2018} also used individuals' information for emotion recognition.
For example, Liu et al.\cite{Liu2020} used a network for recognizing facial emotions and body poses, while Fisher et al~\cite{Yu2019} developed a method to detect facial expressions.
This research goes in a different direction by focusing on global features to avoid harming as far as possible privacy.

The use of synthetic data for model training is a well-established practice in machine learning research.
This data-driven method is often employed by researchers to enhance the generalizability or, a contrario, the specificity of their models.
For instance, Dwibedi et al.~\cite{Dwibedi2020} harnessed real videos to construct synthetic repetition videos for training their ``Counting Out Time'' model.
Marín-Jiménez et al.~\cite{marinjimenez2019laeonet} altered the relative head positions of individuals in images to comply with a real scenario.
Their aim was to improve the generalizability of their Look At Each Other (LAEO) model.
In research specifically related to the EmotiW challenge, Petrova et al.~\cite{Petrova20} employed monomodal synthetic static images to enable their model to concentrate on individuals' faces while disregarding background.
Their method creates static images to train their feature extraction.
This research extends this approach to generate coherent video sequences coupled with class-related sound. 

\section{VGAF Dataset}

The VGAF dataset~\cite{Sharma2019} is a collection of web videos that include a wide variety of genders, ethnicities, event types, numbers of people, and poses.
These videos are grouped into three categories of group level emotions: Positive, Neutral, and Negative (as shown in Figure~\ref{fig:example}).
The classification of each video was determined by a voting process involving several annotators, with a majority vote deciding the final classification.
The dataset is divided into a Training set, a Validation set, and a Test set, with 2661, 766, and 756 videos in each set respectively.
The Validation set is unevenly distributed, with more videos in the Positive and Neutral categories compared to the Negative category.

The videos include groups of at least two people and feature a range of events like interviews, festivals, parties, and protests, among others.
They vary in resolution, and are split up into 5-second labeled clips.
The frame rate varies from 13 to 30 frames per second.
The camera focuses on different positions in each video, resulting in frames that highlight different group organisations.
Thus, the technical  and content varieties of the videos, the association of images and sounds to identify emotions, make the classification complexity of the VGAF dataset. 
 One of the dataset's complexities is that some videos feature children playing and hitting tables or chairs, generating sounds that may be confused with protestation noises.
 However, these two events are different because playing with children creates a positive atmosphere, while protestations generate a negative atmosphere.

\section{Methodology}
\subsection{Model architecture}

\begin{figure*}[ht!]
  \centering
  \includegraphics[width=\linewidth]{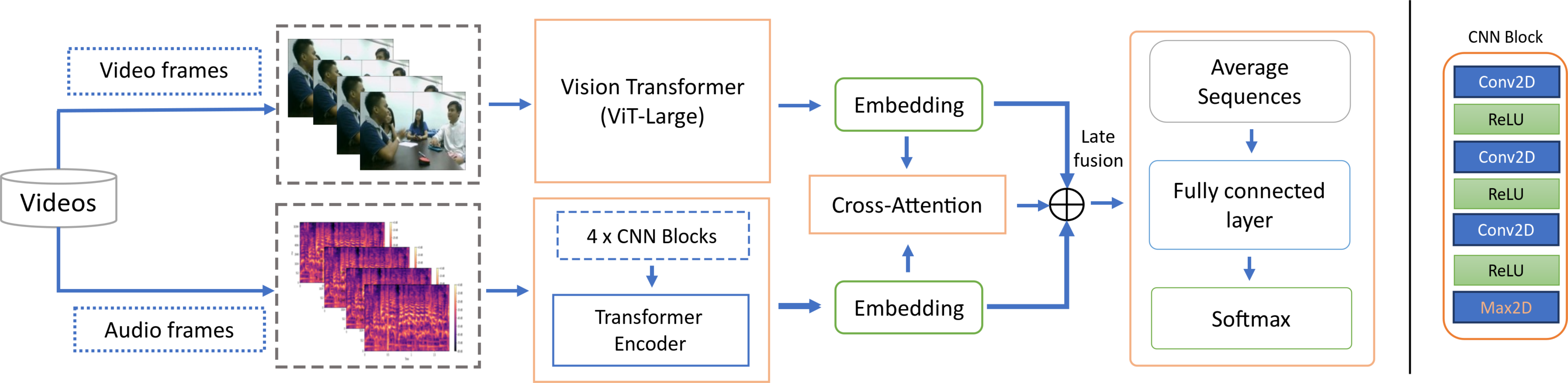}
  \caption{At left, the proposed model is a combination of two monomodal branches, a Cross-Attention and a late fusion paradigm. The video branch uses a pre-trained vision transformer (ViT) model~\cite{Dosovitskiy2020}. The audio branch encompasses 4 CNN blocks followed by a transformer encoder. At right, description of one CNN block used in the audio branch.}
  \label{fig:model}
\end{figure*}

The proposed neural architecture is illustrated in Figure~\ref{fig:model}.
This model encompasses two monomodal branches producing embedding for video and audio frames.
The video branch exploits a pre-trained, then fine tuned, vision transformer (ViT-Large, 14 patches)~\cite{Dosovitskiy2020} ending by a custom linear layer with an output dimension of 1024, i.e. the embedding dimension of the model ($d_{model}$).
The audio branch consumes a sequence of Mel-Spectrograms as input through four specialized CNN blocks.
As depicted, these blocks are composed of 2D convolutions stack with Relu and a final max pooling.
The final section of the audio branch integrates a Transformer Encoder layer, as outlined in~\cite{Vaswani2017} with fixed positional-encoding and four self-attention heads.
The \textit{feed-forward} hidden-size is set at twice the value of $d_{model}$, while output dimension remains $d_{model}$.
To integrate interaction between audio and video modalities, a late fusion mechanism concatenates embeddings from the monomodal branches jointly with a Cross Attention one.
In cross-attention, the audio modality serves as query ($Q$), while the video modality serves as key ($K$) and value ($V$). 
The resulting attention weights are then multiplied with the video embeddings to obtain the final transformed representation.
An average operation is then applied to the concatenated sequence averaging all frame embeddings to one of size $3*d_{model}$.
The classification task is carried out using a fully-connected layer followed by a Softmax activation, which yields to the final classification result in 3 classes.

\subsection{Privacy compliant features}

As stated, one self-imposed constraint is to remain as privacy compliant as possible. 
To do so, our approach carefully avoids all individual features, i.e. features that could identify a person directly or within a group, focusing on global features.

For the video branch, our system does not employ body pose, shape, height, nor body language features like gesture or agitation.
Computed facial information like FACS, emotion or person id are excluded, like counting and tracking individuals within the scene.
The video input is a set of $n$ still video frames uniformly taken over the video and resized to $224\times224$.
$n$ is chosen among two extrema: 1 frame per second (\textit{fps}) and 15 frames per second on the 5-second videos of VGAF.
From our experience, 1 \textit{fps} is a lower band to get a non-random classification.
As 15 \textit{fps} is the lowest \textit{fps} on the training videos, we set it as max \textit{fps} to train on actual frames without repeating some of them.
Thus, $n$ is taken in \{5,75\} for each video in this research.

For the audio processing, we avoid speaker identification, speaker diarization or any speech-to-text processing.
The audio of all videos is standardized by resampling it at 16~Khz and converting it to mono channel.
By compliance with video branch, we extract 5 or 75 audio frames per video.
With 5 frames, the audio frame corresponds to 1 second with no overlap.
In the case of 75 frames, a sliding window is set to 67 milliseconds to get the right number of frames.
Last, each audio frame is converted into Mel-Spectrograms using 128 Mel filters, to produce an input image of $128 \times 251$ for the CNN blocks.

 \subsection{Synthetic video dataset}

Inspired by former research~\cite{Petrova20}, we augmented our training set with synthetic data.
The purpose is to guide the neural network to concentrate on faces while searching for positive, neutral or negative contexts, ignoring the background.
The original generation process involves placing real faces expressing basic emotions~\cite{Ebner2010} onto random backgrounds~\cite{Yu2015} to create still images conveying emotions.
To create synthetic videos, we take from 3 to 9 faces expressing the right emotions, for instance positive emotions to generate positive video, and randomly move them on a fixed background.
To mitigate the impact of occlusions, we introduce masks in the generation process, which allows a maximum of ten percent of occlusion on faces.
As generating audio is a very complex task in our context, we associated randomly an audio with the same class from the VGAF training set to the generated video sequence.
The synthetic video dataset has the same size and class balance as the VGAF training dataset: 802 (30,16\%), 923 (34,71\%), 934 (35,13\%) of positive, neutral and negative videos respectively.

\begin{table}
 \caption{Evaluation of synthetic video ratio on the validation accuracy.}
  \label{tab:syntchoice}
  \vspace{0.5em}
\resizebox{\linewidth}{!}
{
  \begin{tabular}{ccccl}
    \hline
     Synthetic ratio  & Synthetic videos  & Total videos         &  Acc.\\
\hline
 10\%  &297           &2958                 &74.80\%\\
 20\%  &666           &3327                     &{74.93}\%\\
 \textbf{30\%}  & \textbf{1140}          &\textbf{3801}                  &\textbf{75.07}\%\\
 40\%  &1773          &4434                   &74.41\%\\
 50\%  &2661          &5322                  &74.28\%\\
  \hline
\end{tabular}
}
\end{table}

\section{Experiments}

\subsection{Optimal synthetic video ratio}

We conducted experiments to measure the impact of synthetic data ratio on performance using 5 frames per video.
The synthetic ratio was gradually increased from 10\% to 50\% with a 10\% step of the total training data. 
The accuracies on the validation set of these variants can be found in Table~\ref{tab:syntchoice}.
The best accuracy (75.07\%) is reached using 30\% of synthetic data in the training.
Less or more data leads to lower performance.
Following this result, in all subsequent experiments comprising standard and synthetic data, the synthetic ratio is set to 30\%.

\subsection{Multimodal ablation Study}

To evaluate relevance of each of our training inputs, (VGAF video and audio, synthetic data), experiments of all their combinations with 5 and 75 frames were conducted.
Models were trained using cross-entropy loss and the SGD optimizer with a fixed learning rate of $10^{-5}$.
We used the VGAF (audio and/or video) and the synthetic datasets in monomodal and multimodal variants architecture.
The ViT-Large model consists of 24 pre-trained transformer encoder blocks on 14 patches.
Its weights are frozen for 10 epochs, and then released to be fine-tuned on our data.
Cross attention is present only when both video and audio branches of the network are active.
Results on the VGAF validation set are summarized in Table ~\ref{tab:ablation}.

Training on audio data only leads to average performance up to 56.4\% with 5 frames per video.
Using 75 audio frames, the performance decreases.
This may be explained by the averaging effect of the sliding window applied in this case.
Using only synthetic videos performs better with arround 60\% of accuracy, showing  interest of this approach.
The best mono input performance is 74.15\% using 75 video frames.
As in many computer vision tasks, pre-trained vision transformers prove to be efficient to capture relevant features on the VGAF dataset.
Combining audio and videos inputs produces various performance in term of validation accuracy.
Joining synthetic video and audio does not improve much the performance of individual inputs.
The best multimodal performance (78.07\%) is obtained by combining all inputs using 5 frames.
Our best accuracy concerns a monomodal combination of VGAF and synthetic video dataset during training phase.
This model outperforms all other models with a validation accuracy of 79.29\%.

Five monomodal and multimodal versions of our model participated in the EmotiW challenge 2023.
Performances on the test set are reported in Table~\ref{tab:test}.
As anticipated, audio and synthetic data models (v1 and v2) remain around 55\% of test accuracy.
Unlike validation performance, video plus synthetic model (v3) is less accurate than the 2 multimodal ones (v4 and v5).
Noticeably, both multimodal systems, using 5 (v4) or 75 frames (v5) per video, have the same performance even if their predictions differ: their prediction agreement, i.e. when the same class is predicted for the same video, is 88\%.
As expected, prediction agreements of v5 versus v1 and v3 are respectively 53\% and 90\%, highlighting that, for our architecture, audio provides less information than video.

\begin{table}[ht]
\centering

\caption{Ablation study investigating the impact of different inputs on validation accuracy using 5 and 75 frames per video. The ablation considers all combinations of (VGAF) video, (VGAF) audio, and generated synthetic videos.}
\label{tab:ablation}
\vspace{0.5em}
\resizebox{\linewidth}{!}
{

\begin{tabular}{clccc}
\hline
&  \textbf{Input Data}      & \multicolumn{1}{c}{\textbf{5 Frames}} & \multicolumn{1}{c}{\textbf{75 Frames}} \\
\hline 
& audio                          & 56.40\%                       & 54.96\% \\
\hline
\multirow{3}[2]{*}{\begin{sideways}Video\end{sideways}} & synt\_video                  & 57.44\%                       & 60.05\%\\
& video                         & 73.62\%                     & 74.15\%\\
& video + synt\_video             & 75.98\%          & \textbf{79.24}\%\\ 
\hline
\multirow{3}[2]{*}{\begin{sideways}\parbox{4em}{\centering{}Audio\\\centering{}\vspace{-0.5em}+\\\centering{}\vspace{-0.5em}video}\end{sideways}} & synt\_video + audio            & 62.14\%                      & 58.75\%\\             
& video + audio                   & {76.11}\%                      & 77.42\%\\             
& video + audio + synt\_video    & \textbf{78.07}\%                  & {78.72}\%\\ 
\hline

\end{tabular}
}
\end{table}

\begin{table}

\caption{Test Set Accuracy of 5 versions of our model.} %
\label{tab:test}
\vspace{0.5em}
\resizebox{\linewidth}{!}
{
\begin{tabular}{clcc}
\hline
\textbf{Vers.} & \multicolumn{1}{c}{\textbf{Input data}} & \textbf{Nb Frames} & \textbf{Acc.} \\
\hline
v1    & audio & 5     & 55.29\% \\
v2    & synt\_video & 75    & 54.23\% \\
v3    & video + synt\_video & 75    & 74.73\% \\
v4    & video + synt\_video + audio & 5     & 75.13\% \\
v5    & video + synt\_video + audio & 75    & 75.13\% \\
\hline
\end{tabular}%
}
\end{table}

\begin{table}

  \caption{Comparison with SOTA systems on the VGAF dataset. Columns detail usage of individual features, accuracy on official validation set and, when reported, on the test set. The table is ordered by input modalities and then by validation accuracy.}
  \label{tab:comparison}
  \vspace{0.5em}
\resizebox{\linewidth}{!}
{
\begin{tabular}{clccc}
\hline
    &          & \multicolumn{1}{p{4em}}{\textbf{Ind. feat.}} & \textbf{Val. Acc.} & \textbf{ Test Acc. } \\
\hline
\multirow{2}[2]{*}{\begin{sideways}Audio\end{sideways}} & \textit{Ours v1} &        & \textit{56.40\% }&  \textit{55.29\%} \\

      & Ottl et al.~\cite{ottl2020group}  &        & 59.40\% & 62.30\% \vspace{0.2em}\\

\hline
\multirow{4}[2]{*}{\begin{sideways}Video\end{sideways}} & Petrova et al.~\cite{Petrova20}  &        & 52.36\% & 59.13\% \\
      & \textit{Ours v2} &        & \textit{60,05\%} &  \textit{54.23\%}  \\
      & Savchneko et al.~\cite{savchenko2022neural}  &  {\checkmark}   & 70.23\% &  -  \\
      & \textit{Ours v3} &         & \textit{\textbf{79.24\%}} & \textit{74.73\%} \\
\hline
\multicolumn{1}{c}{\multirow{8}[2]{*}{\begin{sideways}Audio + Video\end{sideways}}} & Evtodienko et al~\cite{Evtodienko2021}  &       & 60.37\% & - \\
      & Sharma et al~\cite{sharma2021audio}  &  \checkmark   & 61.61\% &  66.00\% \\
      & Pinto et al.~\cite{pinto2020audiovisual}  &       & 65.74\% & - \\
      & Wang et al.~\cite{wang2020}  &       & 66.19\% &  66.40\% \\
      & Sun et al.~\cite{Sun2020}  &  \checkmark   & 71.93\% & - \\
      & Belova et al.~\cite{belova2022group}  &       & 71.95\% & - \\
      & Liu et al.~\cite{Liu2020}  &  \checkmark   & 74.28\% &  \textbf{76.85\%} \\
      & \textit{Ours v4} &         & \textit{78.07\%} & \textit{75.13\%} \\
      & \textit{Ours v5} &         & \textit{78.72\%} & \textit{75.13\%} \\
\hline
\end{tabular}%
}
\end{table}

\section{Discussion}

Excluding systems from the EmotiW 2023~\cite{emotiw2023} challenge as results were not available at the writing time of this article, one can compare the 5 versions of our model with the state-of-the-art.
Accuracies on validation and test sets are reported in Table~\ref{tab:comparison}.
Using only audio, as we limited input to Mel-spectrograms, our v1 model do not reach score of Ottl et al.~\cite{ottl2020group} proposal, due to their more complex usage of OpenSMILE  features~\cite{eyben2010opensmile} coupled with Deep Spectrum analysis.
On video only system, the v3 model have the best performance even if, as said, its test accuracy is lower than its validation counterpart.
On multimodal systems, v4 and v5 remain under the proposal of Liu et al.~\cite{Liu2020}.
Nevertheless, our approach exposes fair performance while using only global thus privacy-compliant features.

Looking at intrinsic analysis of our models, they benefit from the pre-trained ViT network fine-tuned on the VGAF data and from our synthetic video approach but lack in the audio branch.
Increasing the number of parameters like number of attention heads wages unconditionally to overfitting, showing that such architectures require a lot of training data.
Gathering more group emotion data, (transfer) learning on close or generated data can be part of the solution.
Our synthetic generation process is straightforward and can be enhanced in several ways.
Generative Adversarial Networks can be trained to create new data.
But all these approaches are limited by the complexity and diversity of group emotion videos in terms of audio and video content. 

As anticipated, avoiding individual features leads to set a thread-off between performance and privacy.
In this research, our aim was to investigate the maximum performance that could be obtained without contravening this rule.
Depending on the target application context, some additional information could be added to improve performance.

\section{Conclusion}

This research introduces a privacy-compliant yet efficient method for recognizing group emotions in uncontrolled environments, proposing to eliminate the need for individual feature extraction.
The privacy compliance is achieved through a selection of non-individual characteristics in signal inputs, providing only global information on the scene.
The proposed model is composed of a video and an audio branches.
The video branch is based on a ViT architecture fine tuned to compute relevant embedding on image sequences.
The audio branch extracts Mel-spectrograms and feed them through CNN blocks into a transformer encoder.
A cross-attention mechanism is added before the late multimodal fusion and the final classification.
Our training paradigm includes a generated synthetic dataset to increase the sensitivity of our model on facial expression within the image in a data-driven way.

The extensive experiments show the relevance of our methodology. 
Our proposal performs fairly on the EmotiW challenge, with 79.24\% and 75.13\% of accuracy respectively on validation and test set for the best models.
Contributions of each modality differ, the audio branch having rooms for larger improvement.
Noticeably, our findings highlight that it is possible to reach  75\% of accuracy with privacy-compliant features using only 5 frames uniformly distributed per video.

To conclude, the targeted privacy compliance of this research comes at a cost in performance.
The accuracy remains slightly lower than other methods employing individual features.
The next research challenge is thus to validate whatever it is possible, with individual features, to minimize the performance gap of privacy-compliant models, allowing to use them in more application contexts. 

\section*{ACKNOWLEDGMENTS}
This work was supported by the PERSYVAL Labex (ANR-11-LABX-0025).
This work was granted access to the HPC resources of IDRIS under the allocation 2023-AD010614233 made by GENCI.
We express our gratitude to Garance Dupont-Ciabrini for her contribution to the first step of the synthetic video preparation.

\bibliographystyle{plainnat}
\bibliography{References.bib}

\end{document}